\documentclass[a4paper]{IEEEtran}

\usepackage{cite}
\usepackage{amsmath,amssymb,amsfonts}
\usepackage{algorithmic}
\usepackage{graphicx}
\usepackage{textcomp}
\usepackage{xcolor}
\usepackage{booktabs}

\begin{document}

\title{Enigme: Generative Text Puzzles for Evaluating Reasoning in Language Models}
\author{\IEEEauthorblockN{John Hawkins}\\
\IEEEauthorblockA{\textit{Centre for Artificial Intelligence and Innovation} \\
\textit{Pingla Institute, Sydney, Australia} \\
john@getting-data-science-done.com}
}

\maketitle

\begin{abstract}
Transformer-decoder language models are a core innovation in text based generative 
artificial intelligence. These models are being deployed as general-purpose
intelligence systems in many applications. Central to their utility is the capacity
to understand natural language commands and exploit the reasoning embedded in human 
text corpora to apply some form of reasoning process to a wide variety of novel tasks. 
To understand the limitations of this approach to generating reasoning we 
argue that we need to consider the architectural constraints of these systems.
Consideration of the latent variable structure of transformer-decoder
models allows us to design reasoning tasks that should probe the boundary of
their capacity to reason. We present enigme, an open-source library for generating
text-based puzzles to be used in training and evaluating reasoning skills
within transformer-decoder models and future AI architectures. 
\end{abstract}


\section{Introduction}

The general capabilities of transformer-decoder based large language models has
inspired a range of research into prompt and response computer interfaces
that are being built for a wide range of language processing and creative
purposes\cite{NEURIPS2020_1457c0d6}. 
In addition to focus on text generation, there has been a growing
investigation of the idea that in the process of learning to produce text in a
way that resembles human output, these machine architectures have also learned
to be capable of reasoning\cite{NEURIPS2022_8bb0d291}. 
In addition there have been investigations into whether these models exhibit
\textit{emergent} properties \cite{wei2022emergent} or `sparks
of Artificial General Intelligence` \cite{bubeck2023sparksartificialgeneralintelligence}.
What these investigations all have in common is an attempt to determine
if these large-scale generative models possess a type of intelligence not exhibited
in all of the previous developments of machine learning: general
purpose abstract reasoning that can be applied to new domains.

Research on transformer-based language models often advances
with claims focused on positive examples of a certain behaviour. For example, 
giving the model a set of tasks and reporting on the success at the new task. 
This approach has the advantage of
being able to demonstrate what can be done with these new tools. However, it
suffers from two general problems as a scientific method. The first is that unless
it is accompanied with a clear demonstration that the examples were not in the
training data (either foundation model text or in the fine-tuning process) then
it cannot be known for certain if the behaviour is emergent or memorisation.
This fact has been acknowledged by many of those engaged in studying the emergent
capabilities of language models\cite{bubeck2023sparksartificialgeneralintelligence},
but it is far from being solved methodologically. 
Recent research has begun to shed light on just how much of the output from
large language models is due to the model using its over-parameterisation to
memorize its training data\cite{tirumala2022,hartmann2023}, observations of this
problem have been made on closed-source industry models\cite{HoraceHe2023}. 
New evaluation methods include designing explicit tests of whether language models have a
sufficiently rich model of the world \cite{ivanova2024elementsworldknowledgeewok}, 
which helps filter out when models are merely repeating memorised responses.

Multiple authors have observed that the poverty of our AI benchmarks results
in a constant adjustment of whether a given task requires intelligence or not
\cite{McCorduck2004}. This may be in part because our benchmark driven evaluation means 
that we place emphasis on performance alone, rather than the methodology
used by an algorithm to achieve results\cite{chollet2019measureintelligence}.
Which is further supported by the trend over time in which we re-evaluate our 
evaluation methods in part by understanding the mechanisms by which machine learning
systems seem to game them.
It is worth noting that Large Language Models (LLMs) are routinely evaluated by a
wide variety of criteria\cite{guo2023evaluatinglargelanguagemodels,10.1145/3641289}, 
of which reasoning capability is only one part, and not at the top of the list of 
considerations. Furthermore, there are a wide range of potential problems with 
these models beyond limitations of reasoning 
ability\cite{kaddour2023challengesapplicationslargelanguage}.
Regardless, evaluative focus on the claims of emergent reasoning of these systems 
is important for both the theoretical implications, and due to the 
emerging widespread use of LLMs in machine learning systems 
(e.g. recommendation engines\cite{Wang_Chu_Ouyang_Wang_Hao_Shen_Gu_Xue_Zhang_Cui_Li_Zhou_Li_2024}).
Particularly in the development of agent-based systems\cite{andreas-2022-language}, 
where reasoning ability is a central focus of research\cite{yao2023react}.

Often reasoning is extracted from language models through example-based prompting, 
such as few-shot or chain-of-thought prompting\cite{Wei2022}, which have proven
useful in general as well as domain specific tasks like 
medical reasoning\cite{gramopadhye2024}. These strategies encourage the model
to generate a sequence of tokens that explicitly solve the probelm, but in some
strong sense relies on the user being able to generate a prompt with the
appropriate examples. This requirement demands that we know something about the
structure of the solution
to a problem before asking a language model to solve it, which is only slightly
mitigated by solutions that automate the retrieval of examples\cite{zhang2023automatic}.
Recent research suggests that
multiple language models can potentially overcome this limitation by working in
concert\cite{wang2024rethinkingboundsllmreasoning}.

\begin{figure}
 \includegraphics[scale=0.47]{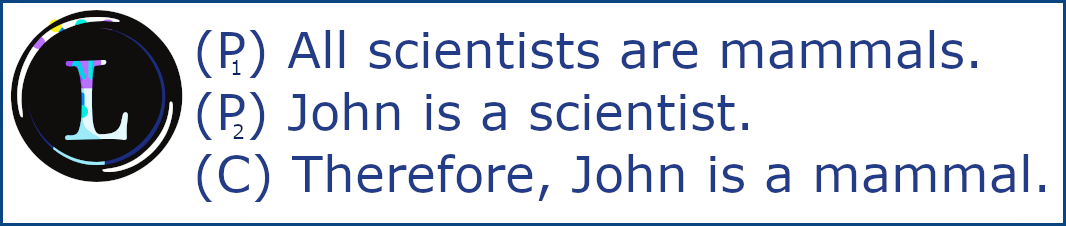}
 \caption{Example of the natural language template for deductive logical reasoning. The conclusion (C) is drawn from the combination of the content in the two premises (P1) and (P2).}
 \label{fig:logic}
\end{figure}

The evaluation of logical reasoning can be broken into different modes, or 
patterns of reasoning, for which different language models exhibit
strengths and weaknesses.
These reasoning modes include strict inference like logical deduction, or 
mathematical reasoning, as well
as more open-ended modes like monotonicity reasoning\cite{yanaka-etal-2019-neural},
and hypothesis generation modes like induction or abduction\cite{he2024idea}.
It has been shown that
performance on these tasks can be broken down into multiple failure modes
and suggested that the variation in performance across models might be due
to the fine tuning process emphasizing certain skills over 
others\cite{xu2023largelanguagemodelsreally}.
These logical reasoning tasks are framed in the most ideal
way for LLMs to succeed, they are often expressed as grammatical language
patterns (as shown in Figure \ref{fig:logic}) that are exactly the kind of
fuzzy template that we expect the latent variable structure of Transformers
to learn\cite{jiang2024peektokenbiaslarge,mirzadeh2024gsmsymbolic}. 
Even on these ideal tasks, it appears that performance can depend 
on arbitrary factors like the order in which premises are 
presented\cite{chen2024premiseordermattersreasoning}, 
supporting the idea that what is happening is
template matching as opposed to genuine reasoning\cite{jiang2024peektokenbiaslarge}.

Abstract reasoning tasks, like those shown in IQ tests, form the basis of
other benchmark datasets (like the ARC challenge\cite{chollet2019measureintelligence}). 
These tasks typically
involve recognising a pattern shown in a sequence of images and then applying 
that pattern to new data, usually under assumptions of perspicacity. When these
visual tasks are converted into text -based patterns, we find that language models
tend to perform poorly, even when fine-tuned for these tasks\cite{gendron2024}.
The abstract reasoning benchmark tasks are a more general form of abductive reasoning
test, the goal is to infer a minimalistic explanation of a small number of observations,
and then successfully apply the inferred pattern. As such they are testing the ability
of a model to derive a world model from impoverished data. We are argue that this
idea can be extended further into an array of tasks that test different dimensions
of the world model inference component of successful reasoning.


\subsection{Motivation}

Our primary concern in this work is to design a system for generating reasoning
puzzles of different forms that can challenge the capabilities of 
transformer-decoder based large language models. 
By considering both the known capabilities of these models and the structure of 
the latent variable space we devise multiple hypotheses about where their
limitations may lie. For each of these hypotheses we design a procedural
generation process to create reasoning puzzles that require abstraction, 
internal model building and alternative interpretations of the symbolic data they
have been trained on. We do this such that the text characters are
arranged in a way that the patterns they signify are difficult to see
without considering them as occupying a physical space that can change over time.
These tasks may require consideration of how the characters appear in 2 dimensions, 
or an understanding of how the text can be used as a 
symbolic illustrations of patterns of processes in the physical world 
that transcend linear arrangements of text tokens.

The rationale for this approach is that the multi-headed attention mechanism makes use
of two avenues for semantic encoding. The first is that tokens exert influence over each
other in a parallel and iterative fashion that allows an embedding to gradually come to
represent the role of a word in a text block. Secondly, in the standard architecture
the system makes use of some
form of positional embedding that characterises the sequential nature of text data.

Given the architectural fundamentals outlined above we design the text puzzles such 
that pure sequential analysis makes the task very difficult. The patterns require
visuualisation across 2 dimensions, and in some cases require abstraction
by considering the text as slices of a 3d pattern.  In addition, we add 
multiple dimensions of variation in these tasks such that the number of potential 
variations is so large that memorisation will be difficult.

Some of the tasks generated in our method bare some similarity with the 
tasks inside the ARC challenge, being composed of sequences of geometric 
(largely grid-based) structures. However, there are several key differences.
(1) Our problems are ASCII character-based puzzles that enable them to be
processed by text only language models. We do this because it removes the
complexity of needing to process the image data and makes the tasks
amenable to pure text based language models, and hence less resource greedy
artificial intelligence research programs.
(2) Our tasks are all built around pattern recognition and reasoning tasks
that should emulate simple world-model building, the formation of abstractions that
are manipulated mentally and intuitive physics reasoning.
We do this because our focus is on fundamentally pragmatic reasoning grounded
in pattern recognition and reasoning about the physical world. These tasks are often
based on a small number of examples and thereby involve a kind of 
inductive bias\cite{goyal2022inductivebiasesdeeplearning} that appears in human reasoning.
In addition, we would like to avoid the potential biases
levelled against abstractions in IQ testing.

\section{Methodology}

In order to provide analytical variation, we define three independent classes of puzzle.
The first class involves \textit{numeric} reasoning tasks, where the text itself is used as a source
for numeric quantities that must be manipulated by the reasoner. The second class are \textit{sequence}
puzzles (similar to common IQ tests) where the reasoner must use abductive reasoning to 
infer the underlying pattern. The final class of \textit{physics} puzzles extends the abductive
reasoning in sequence puzzles to include sequences with behaviors resembling intuitive physics, like
momentum or collision. 
Within each of these classes we then provide additional variation through the complexity
or dimensionality of the puzzles. Finally, each puzzle generator is provided parameterisation
and randomisation to allow for large numbers of variants. 

The total number of potential patterns is determined by the number of free parameters
and varies with each of the puzzle classes. We show estimates of these variation counts
in Table \ref{tab:puzzles}. Note, that as the number of permutations for some of these
puzzles is extremely large, in many instances we are using the largest
possible configuration for the estimate. This number tends to dwarf the smaller configurations
and forms a good lower bound approximation.

\begin{table}
  \caption{Enigme Puzzle Categories and Estimated Variations}
  \label{tab:puzzles}
\begin{tabular}{|l|l|r|}
\toprule
Category    &Dimensions     &Variations          \\
\midrule
Numeric     &1              &$3.0\times 10^{5}$  \\
            &2              &$1.4\times 10^{7}$  \\
            &3              &$4.8\times 10^{9}$  \\
\midrule
Sequence    &1              &$1.6\times 10^{6}$  \\
            &2              &$2.3\times 10^{62}$ \\
            &3              &$6.9\times 10^{33}$ \\
\midrule
Physics     &1              &$1.6\times 10^{6}$ \\
            &2              &$4.6\times 10^{7}$ \\
            &3              &$2.5\times 10^{6}$ \\
\bottomrule
\end{tabular}
\end{table}

\subsection{Numeric Puzzles}

\begin{figure*}
\centering
 \includegraphics[scale=0.45]{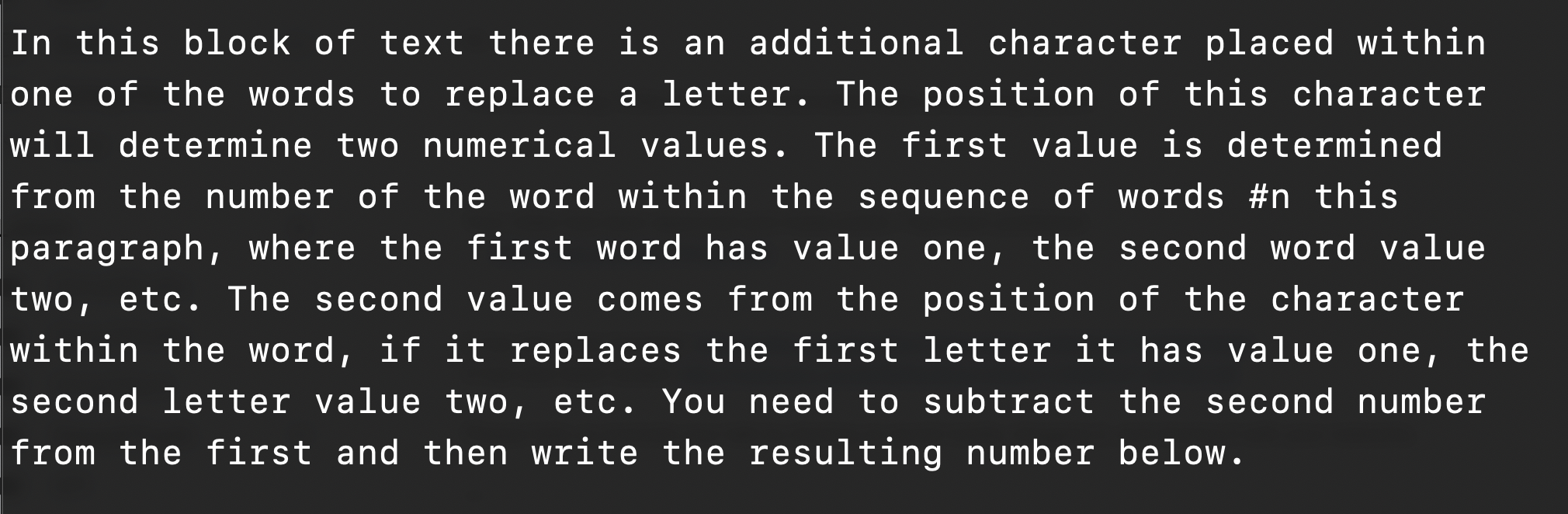}
 \caption{Example 1: Enigme Numeric Puzzle - Self Referential Numeric Puzzle}
 \label{fig:numeric}
\end{figure*}

The numeric puzzles are formed by assembling a final text from a set of 
pre-determined blocks that form multiple variations of the same set of 
instructions. These instructions are self-referential, in that the task they
describe involves analysing the block of text itself.
 
In the generation process we will have replaced some of the characters in the 
words of the instruction block. The instructions require that the agent solving 
the problem numerically analyse the position of these substituted characters
within both the sequence of words and the sequence of letters in the word.
In addition, for the final complexity level, the agent solving the puzzle needs
to recognise the identity of the substituted letter and its position in the sequence
of the alphabet. These are the three potential dimensions to the numeric puzzles:

\begin{itemize}
\item Position in sequence of words
\item Position character within word
\item Position of missing character in alphabet
\end{itemize}
 
These three numeric components are combined in a simple arithmetic operation that
the agent needs to complete to provide the answer. An example of these numeric 
puzzles is shown in \ref{fig:numeric}.

\subsection{Sequence Puzzles}

The algorithm is initialised with a set of parameters drawn randomly from pre-determined
distributions. We create a data structure representing an object of between 1 and 3 dimenions
(controlled by the complexity parameter). 
We randomly select a background character from the set [\verb|.,_'`|] to fill this structure. We then
iterate over the cells in the structure and determine if the background character be replaced
by one of the foreground pattern characters. 

The foreground characters are drawn from a set N of 38 characters, including upper case letters,
numbers and a variety of punctuation characters. Once the initial structure is determined, we then
randomly determine the way in which it will be modified through the series, depending on the 
dimensionality (complexity) of the puzzle. 

\begin{figure*}
\centering
 \includegraphics[scale=0.47]{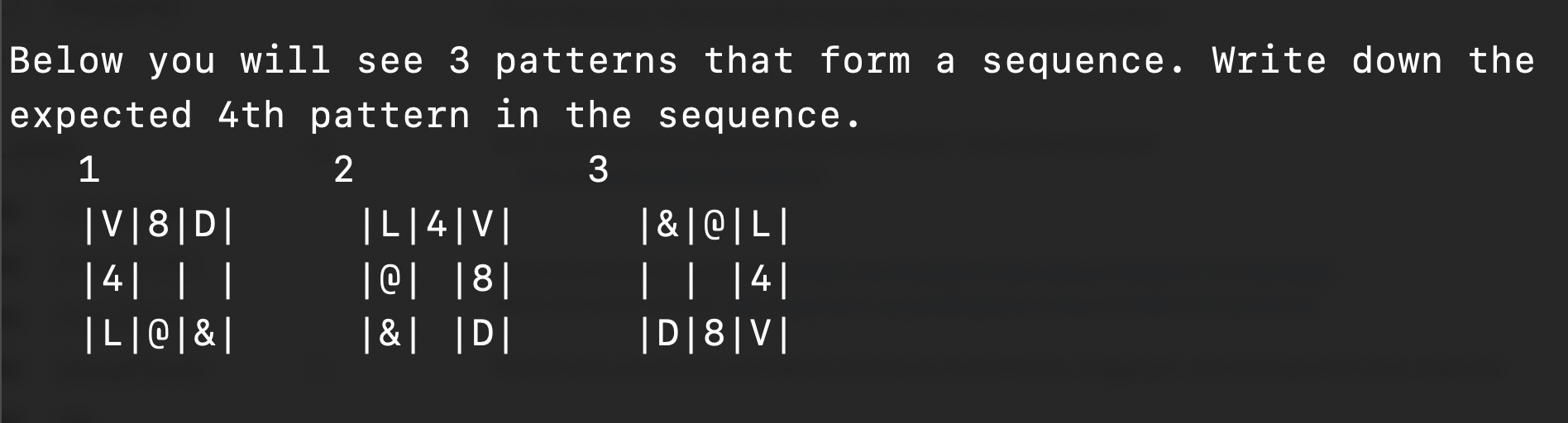}
 \caption{Example 2: Enigme Sequence Puzzle - Patterns Represented with ASCII Text}
 \label{fig:sequence}
\end{figure*}

An example of a puzzle from the sequence class is shown in Figure \ref{fig:sequence}

\subsection{Physics Puzzles}

The physics puzzles share structural similarities with the sequence puzzles;
however, they are designed to emulate the behaviour of very simple physical systems. 
The solution to the puzzle requires extrapolation of that behaviour, invoking notions
of momentum, acceleration, deceleration, or the consequences of objects interacting with
each other and boundaries.
These puzzles demand a form of naive physics reasoning that we hypothesize is essential
to inference of a world model from data.

\begin{figure*}
\centering
 \includegraphics[scale=0.47]{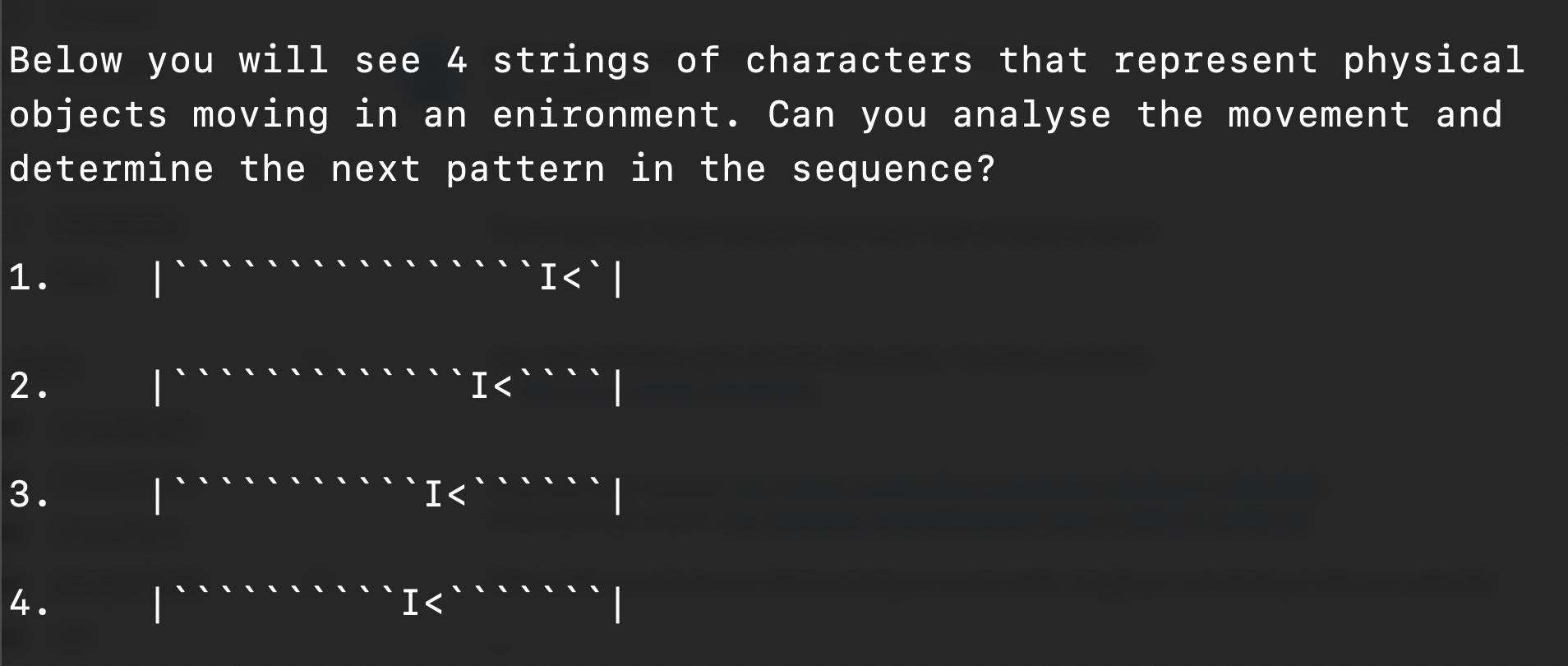}
 \caption{Example 3: Enigme Physics Puzzle - Physical Object Movement Represented with ASCII Text}
 \label{fig:physics}
\end{figure*}

An example of a physics puzzle is shown in Figure \ref{fig:physics}.

\subsection{Data Generation}
 
The \textit{enigme} application can be invoked to generate random puzzles from any of
the three classes. An additional \textit{dimension} parameter determines the complexity
of the puzzles through an abstraction of the idea of problem dimensions. 
We generate one variation from each category and show them in Figures \ref{fig:numeric},
 \ref{fig:sequence}, and \ref{fig:physics}.

These three examples were created by execution of the enigme application using the
following commands respectively:
\begin{verbatim}
  > enigme numeric 2d
  > enigme sequence 2d
  > enigme physics 1d
\end{verbatim}

The puzzles are procedurally generated through a template-based substitution method that allows us
to estimate the total number of potential variations, shown in Table \ref{tab:puzzles}.
The generation process allows strict control over what can be generated,
and programmatic control over the accuracy of the solutions. The estimated number of
these variations reflects the current structure of the package and will increase over
time as more template variations are added.

\section{Conclusion}

We have described three classes of purely text-based reasoning puzzles that have been defined
inside an open-source template-based generative engine. The rationale for these puzzles
was derived from a combination of previous work on different modes of reasoning
and an analysis of the expected limitations of the
latent variable space learned inside the dominant Transformer architecture. We expect this 
package to be extended with additional classes of puzzle as we identify alternative modes
of reasoning that can be tested.

The generation process for these puzzles is parameterised by a dimension/complexity
value, allowing researchers to experiment with training and evaluating
models on their ability to reason on problems of varying difficulty. 
We expect extensive usage of this package for benchmarking the reasoning ability of both
LLMs and future architectures for artificial intelligence systems. Furthermore, we 
anticipate broader potential applications of the enigme package in fields such as 
education and cognitive science.

\section{Availability}

The \textit{enigme} open-source code and python application are available on PyPi
and GitHub (https://github.com/john-hawkins/enigme).

\bibliographystyle{IEEEtran}
\bibliography{IEEEabrv,refs}

\end{document}